\documentclass[conference]{IEEEtran}
\IEEEoverridecommandlockouts
\usepackage[T1]{fontenc}
\usepackage{cite}
\usepackage{amsmath,amssymb,amsfonts}
\usepackage{bbding}
\usepackage{graphicx}
\usepackage{amssymb}
\usepackage{hyperref}
\usepackage{amsmath}
\usepackage{tabularx}
\usepackage{booktabs}  
\usepackage{xcolor}
\usepackage{subcaption}
\usepackage{caption}
\usepackage{mwe}
\usepackage{multirow}
\usepackage[ruled,vlined,linesnumbered]{algorithm2e}

\definecolor{darkgreen}{rgb}{0,0.5,0}

\newcolumntype{C}[1]{>{\centering\arraybackslash}p{#1}}
\newcolumntype{M}{>{\hsize=0.5\hsize\raggedright\arraybackslash}X}
\newcolumntype{Y}{>{\centering\arraybackslash}X}

\def\BibTeX{{\rm B\kern-.05em{\sc i\kern-.025em b}\kern-.08em
    T\kern-.1667em\lower.7ex\hbox{E}\kern-.125emX}}
\begin{document}

\title{Guarding the Middle: Protecting Intermediate Representations in Federated Split Learning}

\author{\IEEEauthorblockN{Obaidullah Zaland\IEEEauthorrefmark{1},
Sajib Mistry\IEEEauthorrefmark{3} , and Monowar Bhuyan\IEEEauthorrefmark{1}}

\IEEEauthorblockA{\IEEEauthorrefmark{1}Department of Computing Science,
        Ume\r{a} Unviersity, Ume\r{a}, SE-90187, Sweden\\
Emails: \{ozaland, monowar\}@cs.umu.se}
\IEEEauthorblockA{\IEEEauthorrefmark{3}Curtin University, Bentley WA 6102, Australia\\
Email:
sajib.mistry@curtin.edu.au}

\thanks{This work was partially supported by the Wallenberg AI, Autonomous Systems and Software Program (WASP) funded by the Knut and Alice Wallenberg Foundation via the WASP NEST project “Intelligent Cloud Robotics for Real-Time Manipulation at Scale.” The computations and data handling essential to our research were enabled by the supercomputing resource Berzelius provided by the National Supercomputer Centre at Linköping University and the gracious support of the Knut and Alice Wallenberg Foundation.}
}

\maketitle

\begin{abstract}
Big data scenarios, where massive, heterogeneous datasets are distributed across clients, demand scalable, privacy-preserving learning methods. Federated learning (FL) enables decentralized training of machine learning (ML) models across clients without data centralization. Decentralized training, however, introduces a computational burden on client devices. U-shaped federated split learning (UFSL) offloads a fraction of the client computation to the server while keeping both data and labels on the clients' side. However, the intermediate representations (i.e., smashed data) shared by clients with the server are prone to exposing clients' private data. To reduce exposure of client data through intermediate data representations, this work proposes k-anonymous differentially private UFSL (KD-UFSL), which leverages privacy-enhancing techniques such as microaggregation and differential privacy to minimize data leakage from the smashed data transferred to the server. We first demonstrate that an adversary can access private client data from intermediate representations via a data-reconstruction attack, and then present a privacy-enhancing solution, KD-UFSL, to mitigate this risk. Our experiments indicate that, alongside increasing the mean squared error between the actual and reconstructed images by up to 50\% in some cases, KD-UFSL also decreases the structural similarity between them by up to 40\% on four benchmarking datasets. More importantly, KD-UFSL improves privacy while preserving the utility of the global model. This highlights its suitability for large-scale big data applications where privacy and utility must be balanced. 

\end{abstract}


\section{Introduction}
Edge applications, where massive and heterogeneous datasets are generated across clients, have increased interest in on-device data-driven decision-making~\cite {joshi2022federated}. Alongside the development of enhanced data privacy and regulatory frameworks~\cite{gdpr}, decentralized machine learning (DML) techniques have garnered significant attention over the past decade. Federated learning (FL) has emerged as a decentralized training paradigm for machine learning (ML) models, where multiple clients collaborate to train a common global model~\cite{mcmahan2017communication,zalandiconip}. Unlike centralized ML training, FL clients retain their data locally, thereby enhancing data privacy and ownership. An alternative DML approach, split learning (SL)~\cite{vepakomma2018splitlearninghealthdistributed}, distributes the model training across a client and a server by model partitioning, while keeping the data at the source. In SL, the initial model layers — and sometimes the last layers — are typically trained on the client (the data source), and training the rest of the model is offloaded to a \textit{more powerful} server, which further highlights its relevance for large-scale big data applications.

Federated split learning (FSL)~\cite{thapa2022splitfed} has recently been explored as a combination of the two decentralized training methodologies. FSL distributes the training across clients and the server in an FL setup to reduce the computational burden on clients while adhering to the integral principle of FL (i.e., local data storage). Two-way federated split learning (TFSL) ~\cite{fedsljiang} divides the training model into two parts. The initial part is trained on the client, while the latter is trained on the server. In a typical TFSL setup, the clients train the initial portion of the model and send the intermediate features (i.e., smashed data)\footnote{In the rest of this paper, smashed data, intermediate client representations, and intermediate features are used interchangeably.} to the server, where the latter portion of the network is trained. The gradients for the latter part are propagated at the server and then sent to the client for completion of the backward step. TFSL, however, suffers from two major issues: i) Clients need to share their labels with the server, which is infeasible under strict data privacy and ownership principles~\cite{fsliot}, and ii) Intermediary features may cause data leakage in the existence of an adversary~\cite{fslprivacy}, who with some knowledge can reconstruct the private client data form the smashed data representations sent to the server.  

\begin{figure*}[!t]
    \centering
    \includegraphics[width=0.9\linewidth]{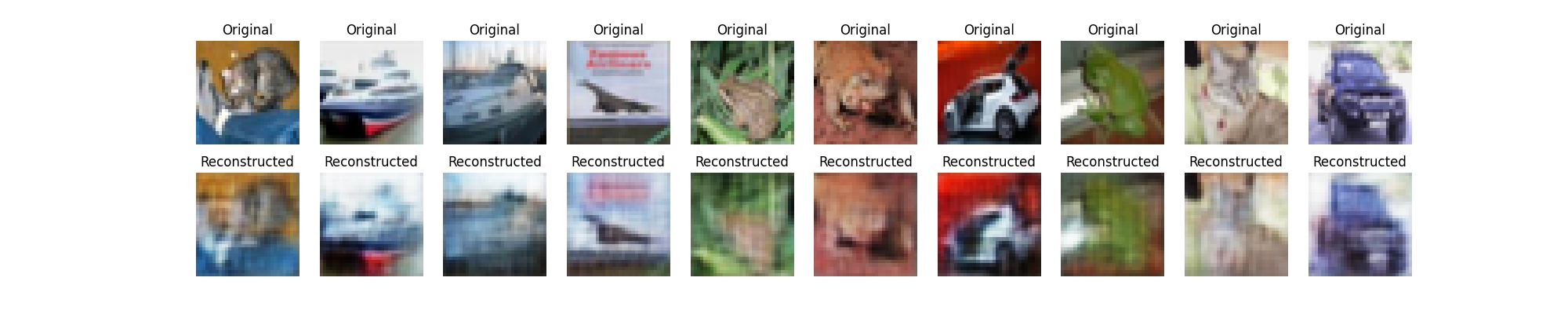}
    \caption{The clients' data on the top, and reconstructed data from the client's smashed data at the bottom.}
    \label{fig:reconstruction}
\end{figure*}
U-shape federated split learning (UFSL)~\cite{vepakomma2018splitlearninghealthdistributed,fesvibs} solves the first problem with TFSL by dividing the training model into three parts, where the first and last portions reside on the client, and the middle part resides at the server. While UFSL solves the label localization problem, it remains susceptible to data leakage from the intermediate \textit{smashed data}~\cite{fslprivacy1d} shared among clients. In the existence of a curious server with very little information, the client's local data can be reconstructed. \textit{Figure} \ref{fig:reconstruction} shows the actual and reconstructed data from the smashed data by a curious server. While there exists minimal literature on privacy for federated split learning, existing studies mostly explore the effects of differential privacy(DP)~\cite{fesvibs} and the choice of the cut layer (the last layer for the client model) to improve smashed data privacy~\cite{fslprivacy}. The choice of cut layer has a significant impact on the division of computational load between the client and server. While a deeper client network might preserve better privacy, it defeats the computationally efficient purpose of FSL for the client. DP, on the other hand, can improve privacy by adding noise to raw data, but can still be susceptible to data reconstruction attacks if $\epsilon$ is higher. The effects of k-anonymity, especially micro-aggregation, have not been explored, either individually or in combination with differential privacy, on privacy in FSL.

This work aims to provide a UFSL approach with enhanced privacy and thus presents k-anonymized differentially private federated U-shaped split learning (KD-UFSL). The main contributions of this work are as follows.
\begin{itemize}
    \item We propose a novel, privacy-enhanced UFSL approach, KD-UFSL, that incorporates data-level differential privacy and feature-level k-anonymity with micro-aggregation to reduce the impact of data reconstruction from the client's smashed data. To our knowledge, this is the first work that applies model-level k-anonymity in federated split learning. Additionally, contrary to most existing works that add DP to the smashed data, KD-UFSL adds noise to the raw data. At the same time, micro-aggregation is applied to the smashed data representations sent by clients.
    \item We show, by designing a data reconstruction adversarial mechanism in the FSL setup, that an adversary can reconstruct raw data from the client's smashed data representations, and then show that KD-UFSL reduces the risk of data reconstruction while preserving the final model utility.   
    \item We provide an experimental evaluation across four different benchmarking datasets, both in RGB and greyscale, to demonstrate the effectiveness of the proposed approach based on the difference between the input and reconstructed data, alongside their structural similarity.
    \item We conduct a utility study of the KD-UFSL framework to demonstrate that not only does KD-UFSL reduce the risk of data reconstruction, but it also maintains similar utility to the vanilla UFSL framework, with a suitable choice of $k$. 
\end{itemize}

\begin{table*}[!t]
    \centering
    
    \begin{tabularx}{0.85\textwidth}{C{1cm}|X|C{1cm}|X}
    \hline
    \textbf{Not.} & \textbf{Explanation} & \textbf{Not.} & \textbf{Explanation}\\
    \hline
    $\mathcal{C}$ & Set of participating clients & $\mathcal{G}$ & Set of client groups \\
    $n$ & Number of participating clients& k & Number of groups for k-anonymity\\
    $\mathbf{D}_c$ & Private dataset at client $c\in\mathcal{C}$& $\mathbf{D}$ & Complete dataset in all clients\\
    $F_c(w)$ & Local objective function for client $c\in\mathcal{C}$ & $w_c^t$ & Local model weights for client $c\in\mathcal{C}$ in iteration $t$\\

    $\mathcal{H}_c$ & Head network for client $c\in\mathcal{C}$& $w_{\mathcal{H}}$ & Weights for the head network\\
    $\mathcal{B}_c$ & Body network for client $c\in\mathcal{C}$& $w_{\mathcal{B}}$ & Weights for the body network\\
    $\mathcal{T}_c$ & Tail network for client $c\in\mathcal{C}$ & $w_{\mathcal{T}}$ & Weights for the tail network\\
    $\mathcal{I}$ & Inversion network & $\eta$ & Learning rate \\

    \hline
    \end{tabularx}
    \caption{Notations used in this paper}
    \label{tab:notations}
\end{table*}

\section{Related Work}

Vanilla federated learning (FL) involves exchanging model weights or gradients between clients and the server, which can potentially cause private client data leakage. Privacy-enhancing methods such as differential privacy (DP) and k-anonymity have thus found increasing applications in FL. DP involves adding noise (usually from a Gaussian or Laplace distribution) to the data before training client models locally~\cite{dpreview}. Yu et al.,~\cite{dpfl} apply DP for content popularity prediction in hierarchical FL settings.  Wei et al.,~\cite{dpfl2} improve privacy by applying DP and sharing partial parameters with the server. Recent methods use adaptive DP to balance the utility-privacy trade-off~\cite{adaptivedp}. K-anonymity~\cite{kann,gupta2025privacy} was developed to anonymize datasets, where each record in the dataset should be indistinguishable from $k-1$ other records. While most of the existing literature in FL focuses on utilizing k-anonymity for anonymizing local datasets~\cite{kann2}, we utilize k-anonymity for smashed data transfers to the server.

The amalgamation of federated learning (FL) and split learning (SL) was first proposed in Splitfed~\cite{thapa2022splitfed}, and has seen an increased interest in various applications, including the Internet of Things (IoT)~\cite{fsliot}, healthcare~\cite{fslhealthcare}, and language modelling~\cite{fsllm}. Although the privacy and security aspects of FL and SL have been explored extensively, the privacy of FSL has received limited attention. On the attack side, prominent attacks in SL, including model inversion attacks (MIAs)~\cite{modelinversion1} and data reconstruction attacks (DRAs)~\cite{DRA2}, can be directly applied to FSL methods. Both MIAs and DRAs can be used to reconstruct the client training data used in the training iteration from the intermediate representations. 

Although SL attacks might be strictly applicable to FSL, the defense mechanism may vary slightly. As FL provides a collaborative mechanism, it presents both opportunities and challenges for improving privacy. PPSFL~\cite{ZHENG2024231} facilitates the protection of model privacy through model decomposition. Yang et al.,~\cite{10256094} incorporated homomorphic encryption (HE) into the aggregation of the client-side model. However, as HE only supports linear operations and is computationally heavy, it introduces its own challenges. Zhang et al.,~\cite{fslprivacy} propose a privacy-aware FSL architecture that incorporates noise addition to the data. The extent of work in FSL privacy has been focused on enhancing either model or data representation privacy individually, especially by adding noise through differential privacy (DP). DP comes with a privacy-utility trade-off~\cite{9069945}. A higher amount of noise can improve privacy, but reduce utility, and vice versa. We argue that a combination of privacy mechanisms applied to both raw data and intermediate representations can enhance privacy.

\section{Preliminaries}

\subsection{Differential Privacy}

\paragraph{Definition 1 ($\epsilon$-DP)~\cite{dwork2014algorithmic}:} A method $M$ is $(\epsilon)$ differentially private if for all neighboring datasets $D_1, D_2$, in the domain of the datasets differing in at most one element, for all $S\subseteq Y$, where $Y$ is the set of all the possible outputs, we have:
\begin{equation}\label{dp}
    Pr[F(D_1)\in S] \leq e^\epsilon Pr[F(D_2)\subseteq S] 
\end{equation}
As the value of epsilon decreases, the privacy guarantee increases. 

\paragraph{Definition 2 ($(\epsilon, \delta)$-DP)~\cite{eddp}:} A method $M$ is $(\epsilon, \delta)$ differentially private if for all neighboring datasets $D_1, D_2$, in the domain of the datasets differing in at most one element, for all $S\subseteq Y$, where $Y$ is the set of all the possible outputs, we have:
\begin{equation}\label{eddp}
    Pr[F(D_1)\in S] \leq e^\epsilon Pr[F(D_2)\subseteq S]  + \delta
\end{equation}
The equation states that the mechanism $M$ is $\epsilon$-DP with probability $1-\delta$.

\paragraph{Defintion 3($(\epsilon, \delta)$-DP~\cite{eddp3}) with Normal Distribution):}
Let $f$ be a function with sensitivity $\Delta f$, and F(x) defined as:

\begin{equation}\label{gaussiandp}
    F(x) = f(x) + \mathcal{N}(0, \sigma^2I)    
\end{equation}
where $\mathcal{N}(0, \sigma^2)$ denotes normal distribution with mean $\mu=0$, and standard deviation $\sigma^2$, then $F(x)$ satisfies $(\epsilon, \delta)-DP$, if 
\begin{equation}\label{normdp}
    \sigma^2 \geq \frac{\Delta f \sqrt{2 \ln(1.25 /\delta }}{\epsilon}
\end{equation}
As $\sigma^2$ is inversely correlated to $\epsilon$, as the value of $\sigma^2$, increases, so does the privacy guarantees.

\section{KD-UFSL: K-Anonymized Differentially Private U-shaped Federated Learning}
Consider a centralized FL setup with a server and a set of $n$ clients $\mathcal{C}$. We refer to Table \ref{tab:notations} for the notations used in this work. In traditional FL, each client $c\in\mathcal{C}$ minimizes a local objective function $F_c(w)$, with respect on its local data $(x^{(i)}, y^{(i)}) \in \mathbf{D_c}$. The global minimization function looks as follows~\cite{varshney2023k}:

\begin{equation}
\min_w \biggl\{F(w) \triangleq \sum_{c=1}^{n} \frac{|\mathbf{D}_c|}{|\mathbf{D}|} F_c(w)\biggl\}
\end{equation}

To achieve the goal mentioned above, in a typical FL setup, in each global training round $t$, the server initiates and distributes an initial global model $w^{t-1}$ to all the participating clients. The clients train the global model and minimize the following loss on their local dataset, to generate local models $w_c^t$, for $c \in \mathcal{C}$, 

\begin{equation}
    \arg\min_{w_c} \sum_i^{|\textbf{D}|}\frac{1}{|\textbf{D}|} l(y^{(i)}, w_c(x^{(i)})) 
\end{equation}
where $l(y^{(i)}, \hat{y}^{(i)})$ is the loss function that calculates the loss between the actual label ($y^{(i)}$) and the predicted label $(\hat{y}^{(i)})$. The clients send their models back to the server, where they are aggregated (e.g., through FedAvg~\cite{mcmahan2017communication}) to form a global model (see \textit{Alg} \ref{alg:fedavg}).

\begin{algorithm}[!t]
\caption{FedAvg}
\label{alg:fedavg}
\SetAlgoLined
\textbf{Server Executes:}\\
Initialize initial weights $w^0$\;
\For{$t \gets 1$ \KwTo $T$}{
    \ForEach{client $c \in \mathcal{C}$}{
        $w_c^t \gets LocalUpdate(c, w_c^{t-1})$\;
    }
    $w^t \gets \sum_{c \in \mathcal{C}} \frac{|\mathbf{D}_c|}{|\mathbf{D}|} w_c^t$\;
}
\textbf{LocalUpdate(c,$w$):}\\
\For{$e \gets 1$ \KwTo $E$}{
    $w \gets w - \eta \nabla \left( \frac{1}{|\mathbf{D}|} \sum_{i=1}^{|\mathbf{D}|} l(y^{(i)}, w(x^{(i)})) \right)$\;
}
\Return{$w$}\;
\end{algorithm}

\begin{figure*}[!t]
    \centering
    \includegraphics[width=0.95\linewidth]{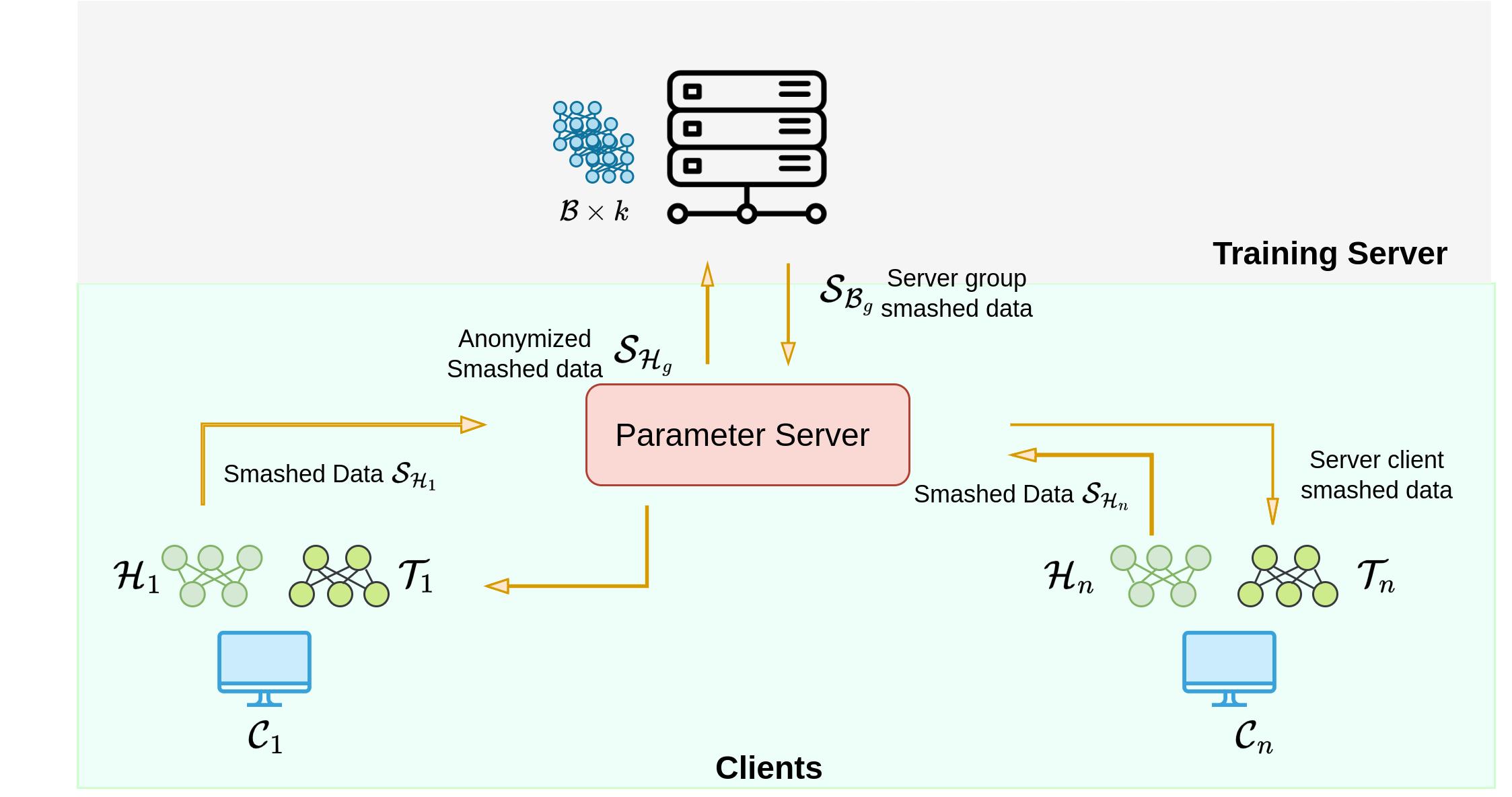}
    \caption{The process of training UFSL with model-level k-anonymity or micro-aggregation. }
    \label{fig:KD-UFSL}
\end{figure*}

In the U-shaped federated split learning (UFSL) settings~\cite{fesvibs}, we divide the local client models into three distinct but connected networks - the head network $\mathcal{H}_c$, the body network $\mathcal{B}_c$, and the tail network $\mathcal{T}_c$, with parameters $w_{\mathcal{H}_c}$, $w_{\mathcal{B}_c}$, and$w_{\mathcal{T}_c}$ respectively - for each client. The head and tail networks for each client reside in the client, while the body network resides in the training server. In this setup, the forward pass is replaced with $\mathcal{T}_c(\mathcal{B}_c(\mathcal{H}_c(x^{(i)})))$. The client takes the input sample batch $x^{(i)}$, and passes it through the head network $\mathcal{H}$, to generate intermediate smashed data $\mathcal{S}_{\mathcal{H}_c}(x^{(i)})$, which is sent to the server, where the body network is utilized to generate server smashed data $\mathcal{S}_{\mathcal{B}_c}(x^{(i)})$, which is sent to the client, and finally passed through the tail network to generate logits. The logits can be converted to predicted class label $\hat{y}^{(i)}$, and the loss can be calculated against the actual class label $y^{(i)}$ (e.g., cross-entropy loss). Essentially, each client minimizes:

\begin{equation}
    \arg\min_{w_{\mathcal{H}_c}, w_{\mathcal{B}_c}, w_{\mathcal{T}_c}} \frac{1}{|\textbf{D}_c|} \sum_{i}^{|\textbf{D}_c|} l(y^{(i)}, w_{\mathcal{T}_c}(w_{\mathcal{B}_c}(w_{\mathcal{H}_c}(x^{(i)}))))
\end{equation}

In UFSL, $n$ clients participate in training their respective head and tail networks, and the \textit{training server} trains $n$ distinct body networks—one for each client. At the end of each global iteration, the head, body, and tail networks are sent to the \textit{parameter server} for aggregation (e.g., using FedAvg). The UFSL setup has two potential points of weakness. The smashed data transmission from the client to the server before the body network processes the input, and the server-to-client transmission after the body network processes the input. In this work, we focus on a federated split learning setup, specifically on the data leakage that might occur from the smashed data sent from the client to the server. Here, we first outline the threat model and then propose a potential solution to mitigate data leakage.

\subsection{Threat Model}
Similar to ~\cite{fslprivacy1d}, we assume an honest, but curious training server. The training server carries out the computation over the body network normally, but is curious about the raw data at the clients. We assume that the training server knows the architecture of the client head network ($\mathcal{H}_c$), and has access to a dataset $\mathbf{D_\mathcal{A}}$ with features similar to the client's private dataset $\mathbf{D_c}$. As the server has access to the intermediary smashed data from the clients' head networks, it can then train an inversion network $\mathcal{I}$. The network $\mathcal{I}$ is trained to convert the smashed representation of a sample batch ($\mathcal{S}_{\mathcal{H}_c}(x^{(i)})$) to the sample batch $x^{(i)}$. In theory, the training server minimizes the following objective function:

\begin{equation}
    \arg\min_{w_{\mathcal{I}}} (MSE(\mathbf{D}_\mathcal{A}, w_{\mathcal{I}}(w_{\mathcal{H}_c}(\mathbf{D_\mathcal{A}}))))
\end{equation}
where $MSE(.)$ represents the mean squared error loss. In summary, the attacker would train the inversion network to convert the head outputs closer to the actual data.

\subsection{Privacy Framework}

The overall privacy framework in KD-UFSL comprises two distinct components, which are explained as follows.
\subsubsection{Differential Privacy (DP)}DP has seen increasing interest recently in improving data privacy~\cite{dpfl} in federated learning (FL). DP attains privacy by adding noise to the individual data samples. Specifically, DP with a \textit{Gaussian mechanism} adds noise sampled from a normal distribution with mean $\mu$ and standard deviation of $\sigma^2$. For each sample batch of data $x^{(i)}$, the differentially private data $\hat{x}^{(i)}$ is calculated as follows:

\begin{equation}
    \hat{x}^{(i)} = x^{(i)} + \mathcal{N}(\mu, \sigma^2I)
\end{equation}
If the mean is set to 0, then 
\begin{equation}
    \hat{x}^{(i)} = x^{(i)} + \mathcal{N}(0, \sigma^2I)
\end{equation}
here, the new data $\hat{x}^{(i)}$ is $(\epsilon, \delta)$-DP if eqn. \ref{normdp} holds.

\subsubsection{K-Annonymity} 
Once noise is added to the data of each individual client, we apply \textit{model-level k-anonymity} to further enhance privacy. The process, illustrated in \textit{Figure}~\ref{fig:KD-UFSL}, proceeds as follows. In each training epoch, the parameter server partitions the set of clients $\mathcal{C}$ into groups $\mathcal{G} = \{g_1, g_2, \dots, g_m\}$, where each group $g_i$ contains at least $k$ clients, i.e., $|g_i| \geq k$. For a client $c \in g_i$, the forward pass produces smashed head representations $\mathcal{S}_{\mathcal{H}_c} = w_{\mathcal{H}_c}(\hat{x}_c^{(i)})$, where $\hat{x}_c^{(i)}$ is the privatized input after Gaussian perturbation. 

The parameter server then performs microaggregation on the smashed data for each group. Specifically, for a group $g_i$, the aggregated representation is given by:
\begin{equation}
    \mathcal{S}_{\mathcal{H}_{g_i}} = \frac{1}{|g_i|}\sum_{c \in g_i} \mathcal{S}_{\mathcal{H}_c}
\end{equation}
This ensures that the smashed representation of any single client is indistinguishable from at least $k-1$ other clients in the same group, thereby satisfying $k$-anonymity at the model level. 

The micro-aggregated representation $\mathcal{S}_{\mathcal{H}_{g_i}}$ is then passed through the server body network $w_{\mathcal{B}}$ to generate the server smashed data $\mathcal{S}_{\mathcal{B}_{g_i}} = w_{\mathcal{B}}(\mathcal{S}_{\mathcal{H}_{g_i}})$. This output is subsequently distributed back to the clients in group $g_i$, where each client’s tail network $w_{\mathcal{T}_c}$ computes the final prediction $\hat{y}_c = w_{\mathcal{T}_c}(\mathcal{S}_{\mathcal{B}_{g_i}})$. The loss $l(y^{(i)}, \hat{y}_c^{(i)})$ is then computed locally to update the client’s head and tail networks, as well as the group’s body network. 

In this way, the micro-aggregation process ensures that individual client representations are obfuscated within a group of size $k$, making it challenging for an adversary at the server to reconstruct any single client’s data. The two-level privacy mechanism, operating at both the data level and the model level, enhances the overall framework's privacy. Alg. \ref{alg:KD-UFSL} provides a detailed overview of the KD-UFSL framework.

\begin{algorithm}[!t]
\caption{KD-UFSL}
\label{alg:KD-UFSL}
\textit{\textbf{Training Server Executes:}}\\
Initialize and send initial weights $w_{\mathcal{H}}^0, w_{\mathcal{B}}^0, w_{\mathcal{T}}^0$\\
\For{$t=1$ \KwTo $T$}{
    $w_{\mathcal{G}}^t \gets GroupClients(w_{\mathcal{B}_1}^{t-1}, ..., w_{\mathcal{B}_n}^{t-1}, k)$ \tcp*{Group clients in size of k}
    \ForEach{client $c \in \mathcal{C}$}{
        $\hat{x}_c^{(i)} \gets x_c^{(i)} + \mathcal{N}(0, \sigma^2I)$ \tcp*{Add noise to batch} 
        $\mathcal{S}_{\mathcal{H}_c} \gets w_{\mathcal{H}_c}(\hat{x}_c^{(i)})$\;
    }
    \ForEach{group $g \in \mathcal{G}$}{
            $\mathcal{S}_{\mathcal{B}_g} \gets w_{\mathcal{B}_g}(\mathcal{S}_{\mathcal{H}_c})$ \;
    }
    
    \ForEach{client $c \in \mathcal{G}$}{
       $\hat{y} \gets w_{\mathcal{T}_c}(\mathcal{S}_{\mathcal{B}_g})$\; 
       $w_{\mathcal{T}_c}, w_{\mathcal{G}_g}, w_{\mathcal{H}_c} \gets \text{Update based on loss } l(y^{(i)}, \hat{y}^{(i)})$\;
    }
    $w_{\mathcal{H}}^{t} \gets \frac{1}{n}\sum_{c\in\mathcal{C}}w_{\mathcal{H}_c}^t$  \tcp*{aggregate client head networks} 
    $w_{\mathcal{B}}^{t} \gets \frac{1}{k}\sum_{g\in\mathcal{G}}w_{\mathcal{B}_g}^t$ \tcp*{aggregate group body networks} 
    $w_{\mathcal{T}}^{t} \gets \frac{1}{n}\sum_{c\in\mathcal{C}}w_{\mathcal{T}_c}^t$ \tcp*{aggregate client tail networks}   
}
\end{algorithm}


\section{Implementation Details}

\subsection{Datasets}
The experiments in this work are carried out on four benchmarking datasets - CIFAR10\footnote{\url{https://www.cs.toronto.edu/~kriz/cifar.html}}, EMNIST~\cite{cohen2017emnist}, FashionMNIST~\cite{xiao2017fashion}, and SVHN\footnote{\url{http://ufldl.stanford.edu/housenumbers/}}. The details of the datasets are presented in \textit{Table} \ref{tab:datasets}. For all the datasets except the CIFAR10, we selected a subset of the dataset, with different fractions. 

\begin{table}[]
    \centering
    
    \begin{tabularx}{\linewidth}{XXXX}
    \hline

    \hline
         Dataset & No. of samples & No. of classes & Fraction used  \\
         \hline
         
         \hline
         CIFAR10 & 60,000 & 10 & 1.0 \\
         EMNIST & 800,000 & 26 & 0.1 \\
         FashionMNIST & 70,000 & 10 & 0.4 \\
         SVHN & 600,000 & 10 & 0.1 \\
    \hline
    
    \hline
    \end{tabularx}
    \caption{Datasets used in this work.}
    \label{tab:datasets}
\end{table}

\begin{figure}[!h]
    \centering
    \includegraphics[width=\linewidth]{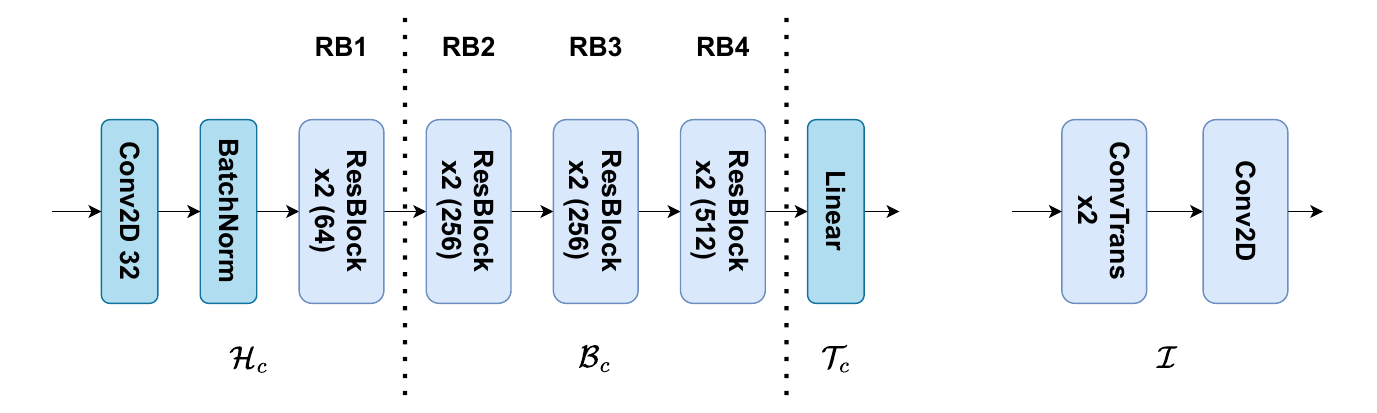}
    \caption{ResNet-18 architecture split into head, body, and tail networks alongside the inversion network. RB1 refers to ResBlock1 and so on.}
    \label{fig:resnet}
\end{figure}
]\begin{table*}[]
    \centering
    \begin{tabularx}{0.8\linewidth}{m{2.1cm}|m{2.5cm}|c|YYYY}
    \hline
    
    \hline
         Network & Method & Metric & CIFAR10 & EMNIST & FashionMNIST & SVHN \\
    \hline
    
    \hline
         \multirow{8}{*}{ResNet18} 
            & \multirow{2}{*}{UFSL} 
                & MSE $\uparrow$ & 0.007 & 0.344 & 0.655 & 0.046 \\
            &   & SSIM $\downarrow$ & 0.949 & 0.159 & 0.212 & 0.846 \\
    \cline{2-7}
            & \multirow{2}{*}{UFSL+DP $_{\sigma^2=0.2}$} 
                & MSE $\uparrow$ & 0.010 & 0.302 & 0.403 & 0.054 \\
            &   & SSIM $\downarrow$ & 0.922 & 0.240 & 0.176 & 0.758 \\
    \cline{2-7}
            & \multirow{2}{*}{UFSL+KA $_{k=3}$} 
                & MSE $\uparrow$ & 0.035 & \textbf{0.417} & \textbf{0.881} & 0.047 \\
            &   & SSIM $\downarrow$ & 0.752 & 0.164 & 0.150 & 0.852 \\
    \cline{2-7}
            
            & \multirow{2}{*}{\textbf{KD-UFSL}} 
                & MSE $\uparrow$ & \textbf{0.086} & 0.312 & 0.428 & \textbf{0.099} \\
            &   & SSIM $\downarrow$ & \textbf{0.547} & \textbf{0.113} & \textbf{0.119} & \textbf{0.572} \\
    \hline
         \multirow{8}{*}{ConvNet} 
            & \multirow{2}{*}{UFSL} 
                & MSE $\uparrow$ & 0.307 & 0.454 & 0.578  & 0.289 \\
            &   & SSIM $\downarrow$ & 0.118 & 0.012 & 0.022 & 0.111 \\
    \cline{2-7}
            & \multirow{2}{*}{UFSL+DP $_{\sigma^2=0.2}$} 
                & MSE $\uparrow$ & \textbf{0.327} & 0.472 & 0.578 &  0.235\\
            &   & SSIM $\downarrow$ & 0.108 & 0.009 & \textbf{0.022} & 0.045 \\
    \cline{2-7}
            & \multirow{2}{*}{UFSL+KA $_{k=3}$} 
                & MSE $\uparrow$ & 0.323 & 0.534 & 0.659 & 0.256 \\
            &   & SSIM $\downarrow$ & 0.034 & 0.008 & 0.060 & 0.163 \\
    \cline{2-7}
            & \multirow{2}{*}{\textbf{KD-UFSL}} 
                & MSE $\uparrow$ & 0.285 & \textbf{0.557} & \textbf{0.676} &\textbf{0.277}  \\
            &   & SSIM $\downarrow$ & \textbf{0.034} & \textbf{0.008} & 0.026  & \textbf{0.161} \\
    \hline
    \multirow{8}{*}{ResNet50} 
            & \multirow{2}{*}{UFSL} 
                & MSE $\uparrow$ & 0.292 & 0.367 & 0.494 &  0.190 \\
            &   & SSIM $\downarrow$ & 0.065 & 0.017 & 0.024 &  0.134\\
    \cline{2-7}
            & \multirow{2}{*}{UFSL+DP $_{\sigma^2=0.2}$} 
                & MSE $\uparrow$ &  0.306 & 0.472 & \textbf{0.733}  &  0.181\\
            &   & SSIM $\downarrow$ &  0.094 & \textbf{0.009} & 0.018 &   0.142\\
    \cline{2-7}
            & \multirow{2}{*}{UFSL+KA $_{k=3}$} 
                & MSE $\uparrow$  & 0.332 & 0.442 & 0.660 &  \textbf{0.295}\\
            &   & SSIM $\downarrow$  & \textbf{0.026} & 0.013 & 0.024 & 0.162 \\
    \cline{2-7}
            & \multirow{2}{*}{\textbf{KD-UFSL}} 
                & MSE $\uparrow$ & \textbf{0.335} & \textbf{0.477} & 0.659 &  0.277\\
            &   & SSIM $\downarrow$  & 0.033 & 0.011 & \textbf{0.007} & \textbf{0.124} \\
    \hline
    
    \hline
    \end{tabularx}
    \caption{Averaged test dataset per image MSE and SSIM between actual and reconstructed data for KD-UFSL and baselines, across different networks. (Bold results represent the best value over the network and the dataset.)}
    \label{tab:results}
\end{table*}
\subsection{Model Training and Hyperparameters}
This work utilizes three different network architectures: ResNet18 and ResNet50 without pre-trained weights, and a lightweight convolutional neural network with four convolution layers. The networks are split into the head $\mathcal{H}$, body $\mathcal{B}$, and tail $\mathcal{T}$ networks. As an example, we show the ResNet18 division in \textit{\textit{Figure}} \ref{fig:resnet}, alongside the inversion network $\mathcal{I}$, which is essentially an inversion of the head network. Unless otherwise specified, the number of clients for all experiments is set to 10. The learning rate for the experiments is set to $0.001$, the batch size is set to $128$, the optimizer is set to Adam, and the standard deviation for DP is set to $0.2$ and $0.3$, respectively. For KD-UFSL, it is set to $0.1$. The value of $k$ is set to \textit{three}, unless stated otherwise. Furthermore, FedAvg is utilized as the global model aggregator for all experiments.

\subsection{Baselines}
Due to the lack of privacy exploration in UFSL, we choose the traditional UFSL setup as the baseline. Essentially, KD-UFSL is evaluated against the following setups:
\begin{itemize}
    \item UFSL: Traditional UFSL setup without any privacy mechanisms. The clients train head and tail networks, and the training server trains the body networks for each client. At the end of each global iteration, the parameter server aggregates all three networks. 
    \item UFSL + DP: UFSL setup where local DP, essentially noise from a normal distribution, is added to the data before training. 
    \item UFSL + KA: UFSL setup, where k-anonymity is applied on the model level during aggregation. 
\end{itemize}

\subsection{Metrics}
The KD-UFSL framework reduces the similarity between the actual data and a potential attacker's reconstructed data from the client's smashed data. As a result, we choose two metrics to evaluate the performance of KD-UFSL against baselines, while for the utility, we compare the accuracy of the final global model.
\paragraph{Mean squared error (MSE):}
The mean squared error calculates the mean squared difference between individual pixels of two images. Given two images $p$ and $q$, MSE is calculated as follows.
\begin{equation}
    MSE(p, q) = \frac{1}{N} \sum_{i=1}^{N} (p_i - q_i)^2
\end{equation}
where N is the total number of pixels in the images. As MSE is the difference between the two pixels, a higher MSE means lower similarity between the two images and vice versa.

\paragraph{Structural similarity index (SSIM)~\cite{nilsson2020understanding}:} SSIM measures structural similarity between two images. Given two images $p$ and $q$, SSIM is calculated as:

\begin{equation}
    SSIM(p,q) = \frac{(2\mu_p \mu_q + C_1)(2\sigma_{pq} + C_2)}{(\mu_p^2 + \mu_q^2 + C_1)(\sigma_p^2 + \sigma_q^2 + C_2)}
\end{equation}
where $\mu_p$ is the mean of image $p$, $\mu_q$ is the mean of image $q$. $\sigma_{pq}$ is the co-variance of the two images, and $\sigma_p^2$ and $\sigma_q^2$ are the variances for images $p$ and $q$ respectively. As SSIM measures similarity, a lower value between the original and reconstructed images is desirable for privacy. $C_1$ and $C_2$ are small real values to prevent division by 0.

\begin{figure*}
    \centering
    \includegraphics[width=0.9\linewidth]{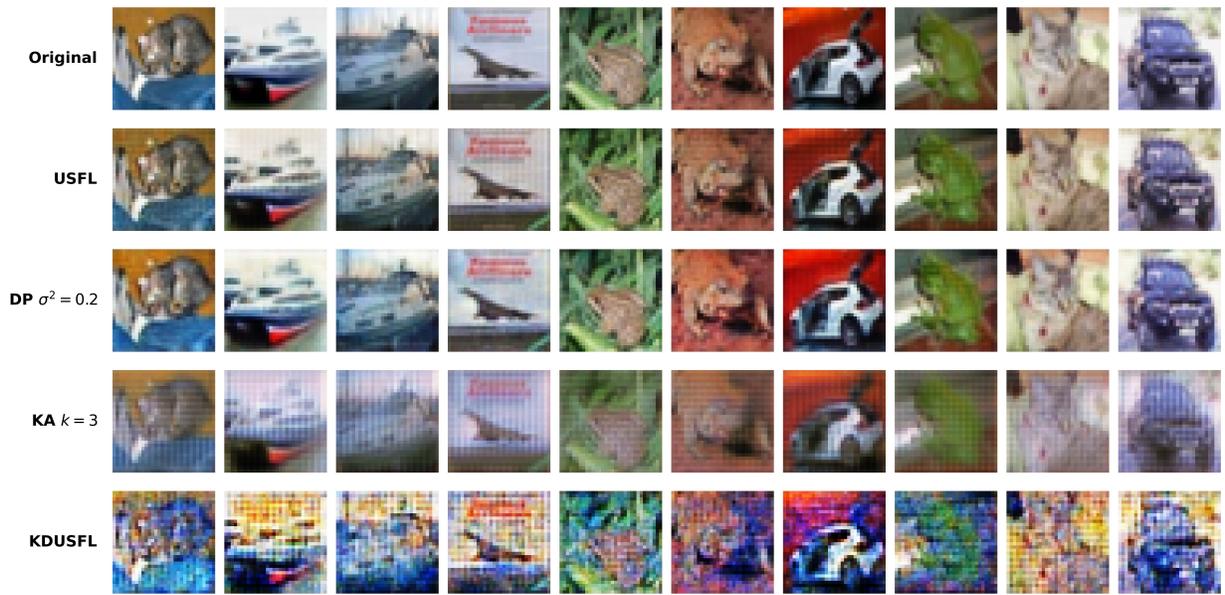}
    \caption{Actual images and reconstructed images from the client's smashed data of different training strategies, using the inversion network. Each row represents a single model.}
    \label{fig:visual}
\end{figure*}

\section{Results and Analysis}
\subsection{Main Results}

The main results presented in \textit{Table} \ref{tab:results} provide a comparison of KD-UFSL against the baselines. We report the average mean squared error (MSE) and the average structural similarity index measure (SSIM) between the actual images and the reconstructed images from the smashed data. The results indicate that KD-UFSL outperforms all the baselines while keeping the least structural similarity between the actual and reconstructed data in most cases. Specifically, the ResNet18 network also performs better than vanilla UFSL and its differentially private counterpart in maximizing the mean squared error. In some scenarios, the gains are more than 50\% compared to the vanilla UFSL setup (MSE on CIFAR10 and SSIM on FashionMNSIT), ensuring that the reconstructed data has less similarity to the actual data. An interesting observation is that k-anonymized UFSL performs similarly to, or in some cases, better than, the differentially private UFSL setup. In two different scenarios, UFSL with KA performs better on MSE loss than KD-UFSL.

In the ConvNet architecture, the results similarly demonstrate KD-UFSL's better performance compared to the baselines, with the highest gains on the SVHN and EMNIST datasets. Specifically, KD-UFSL improves both SSIM and MSE over the SVHN and EMNIST datasets, while achieving a better SSIM value on CIFAR10 and a better MSE value on FashionMNIST. The results for the ResNet50 architecture are somewhat mixed. In ResNet50 experiments, in some cases, either k-annonymity or differential privacy alone performs better than the proposed framework. Overall, the experiments indicate that KD-UFSL outperforms the baselines in most experimental scenarios, as measured by both MSE and SSIM metrics. Specifically, the results on RGB datasets (CIFAR10 and SVHN) show better utility of the proposed framework compared to MNIST-based datasets.

\subsection{Visual Invertibility}
In this section, we show the visual reconstruction of the images. \textit{Figure} \ref{fig:visual} shows the actual and reconstructed images from the clients' smashed data for each model. For each training methodology, the client head model is chosen randomly. Furthermore, to reduce bias, the images are chosen from the test set, ensuring that they were not part of the training for any client or the inversion network. 
\begin{figure*}
    \centering
    \includegraphics[width=\linewidth]{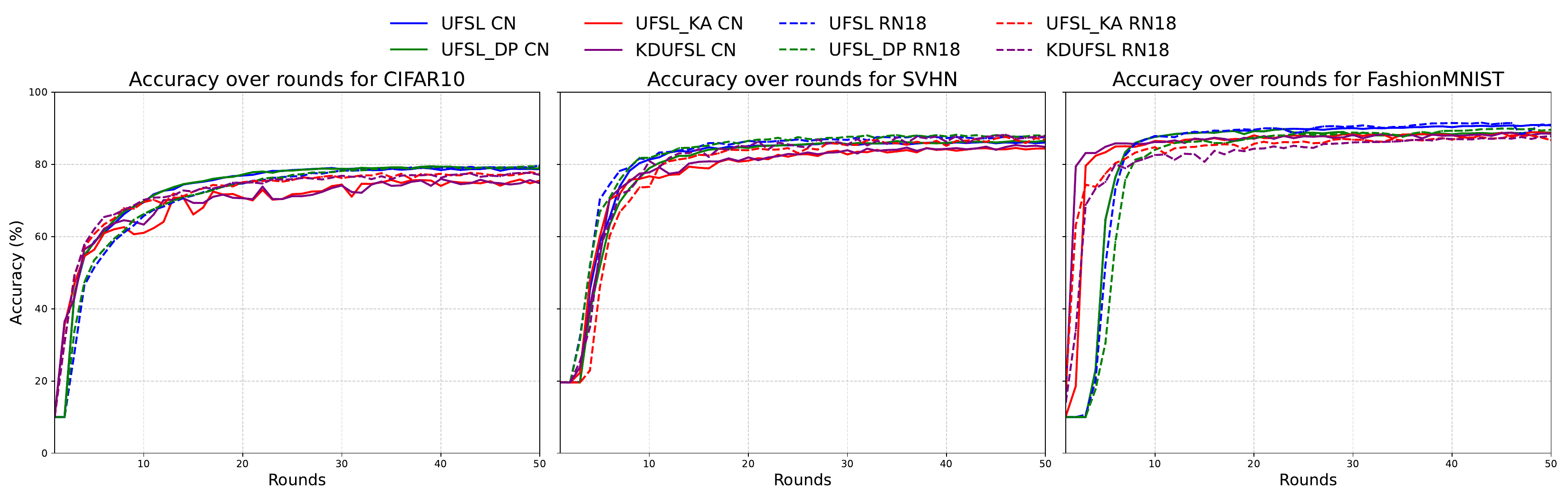}
    \caption{Accuracy over global rounds of the final global model for KD-UFSL and baselines. CN refers to ConvNet, and RN18 refers to ResNet18.}
    \label{fig:utility}
\end{figure*}

\subsection{Utility Discussion}
Apart from enhancing the privacy of the overall framework, KD-UFSL also performs comparably to the vanilla federated split learning (FSL) model in terms of performance. The utility of the final model is within a 2-2.5\% range of the absolute performance of the traditional FSL, and almost on the same lines as UFSL with k-anonymity alone, while improving privacy. \textit{Figure} \ref{fig:utility} illustrates the accuracy of the final model produced over global iterations in KD-UFSL and other baselines on three datasets, comprising a mix of both RGB datasets (CIFAR10 and SVHN) and grey-colored image datasets (FashionMNIST). In both cases, it is visible that the drop in utility with KD-UFSL is minimal when compared to vanilla UFSL-trained global models.




\begin{figure*}
    \centering
    \includegraphics[width=0.95\textwidth]{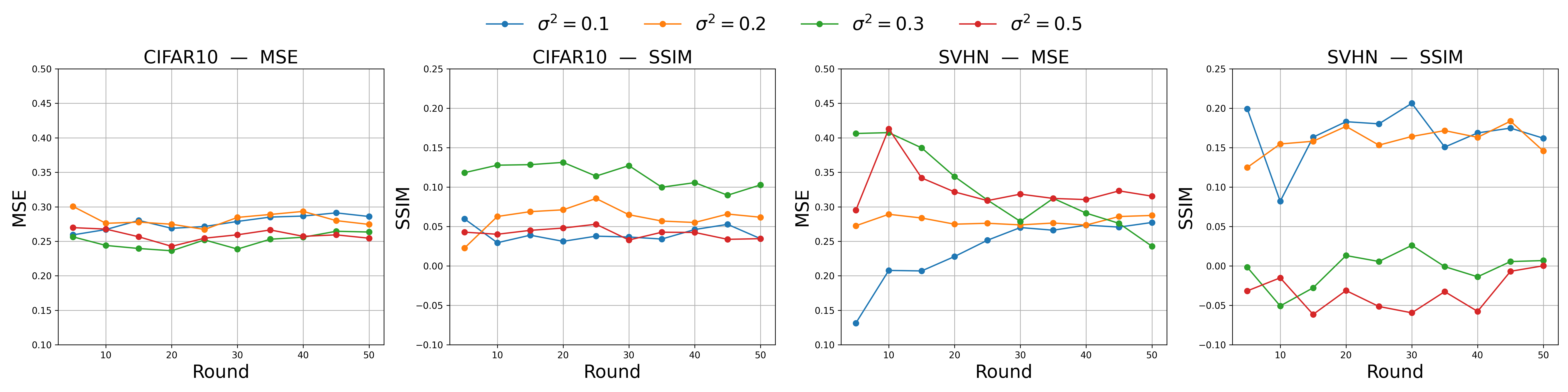}
    \caption{Impact of $\sigma^2$ on MSE and SSIM every 5th global iteration on the ConvNet architecture.}
    \label{fig:sigma}
\end{figure*}

\begin{figure*}
    \centering
    \includegraphics[width=0.95\textwidth]{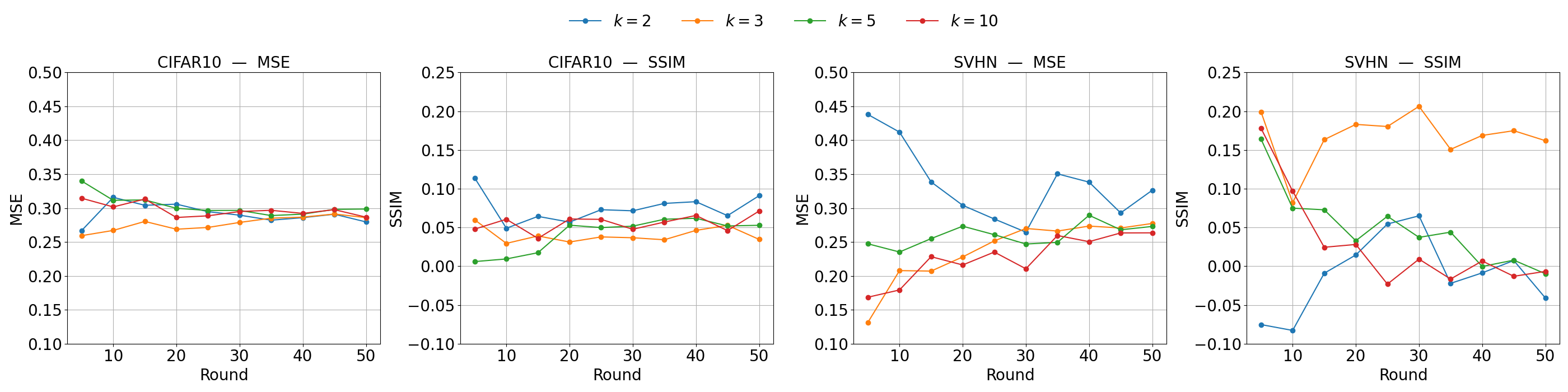}
    \caption{Impact of $k$ on MSE and SSIM every 5th global iteration on the ConvNet architecture.}
    \label{fig:k}
\end{figure*}
\subsection{Impact of $\sigma^2$ and \textit{k}}
The two important hyperparameters in KD-UFSL are the values of $k$ for k-anonymity and the value of $\sigma^2$ for DP. \textit{Figure} \ref{fig:sigma} shows the impact of $\sigma^2$ values on the privacy metrics (SSIM and MSE) of the global model. It can be seen that changing the $\sigma^2$ value in the range (0.1 to 0.5), while having minimal impact on the CIFAR10 dataset, improves both the SSIM and MSE scores on the SVHN dataset. Furthermore, the $\sigma^2$ scores of 0.1 and 0.2 are very similar across both datasets, while the scores for $\sigma^2$ values of 0.3 and 0.5 are also comparable.

\textit{Figure} \ref{fig:k}, on the other hand, shows the impact of the value of $k$, or group sizes, on image reconstruction in KD-UFSL. While the increase in $k$ improves the privacy of the overall framework, the difference is mostly negligible, except for the SSIM score on the SVHN dataset, where the $k=3$ performs the worst.

\subsection{Impact of Head Network Depth}
In this section, we investigate the impact of the head network depth on image reconstruction, and specifically focus on the ResNet18 architecture. For the entirety of the experiments above, the architecture presented in \textit{Figure} \ref{fig:resnet} is used for the head, body, tail, and inversion networks. This section explores deepening the head network and its impact on the overall privacy in KD-UFSL. Based on the naming convention in \textit{Figure} \ref{fig:resnet}, the depth of the head network up to and including layers RB1, RB2, and RB3 is investigated. Keep in mind that the architecture of the inversion and body networks also changes as we change the head network architecture.

\textit{Table} \ref{tab:headdepth} provides a comparative study of KD-UFSL performance with varying depths for the head network. While privacy increases as the head network deepens (as confirmed by lower SSIM and higher MSE for depths up to RB3), it also increases the computational burden on the client, as increased training will take place locally. 

\begin{table}[]
    \centering
    \begin{tabularx}{\linewidth}{m{2cm}|XX|XX}
    \hline

    \hline
         Head Depth & \multicolumn{2}{c|}{CIFAR10} & \multicolumn{2}{c}{SVHN}  \\
         \hline
         
         \hline
          & MSE $\uparrow$ & SSIM $\downarrow$ & MSE $\uparrow$ & SSIM $\downarrow$   \\
         \hline
         RB1 &  0.045 & 0.762 & 0.057 &  0.882\\
         \hline
         RB2 & 0.086 & 0.547 & 0.099  & 0.572 \\
         \hline
         RB3 &  0.313 & 0.035 & 0.274   & 0.043 \\
    \hline
    
    \hline
    \end{tabularx}
    \caption{Impact of head network depth on KD-UFSL performance. Depth up to layer RBi also includes layer RBi.}
    \label{tab:headdepth}
\end{table}

\subsection{Number of Clients}
\begin{figure}
    \centering
    \includegraphics[width=\linewidth]{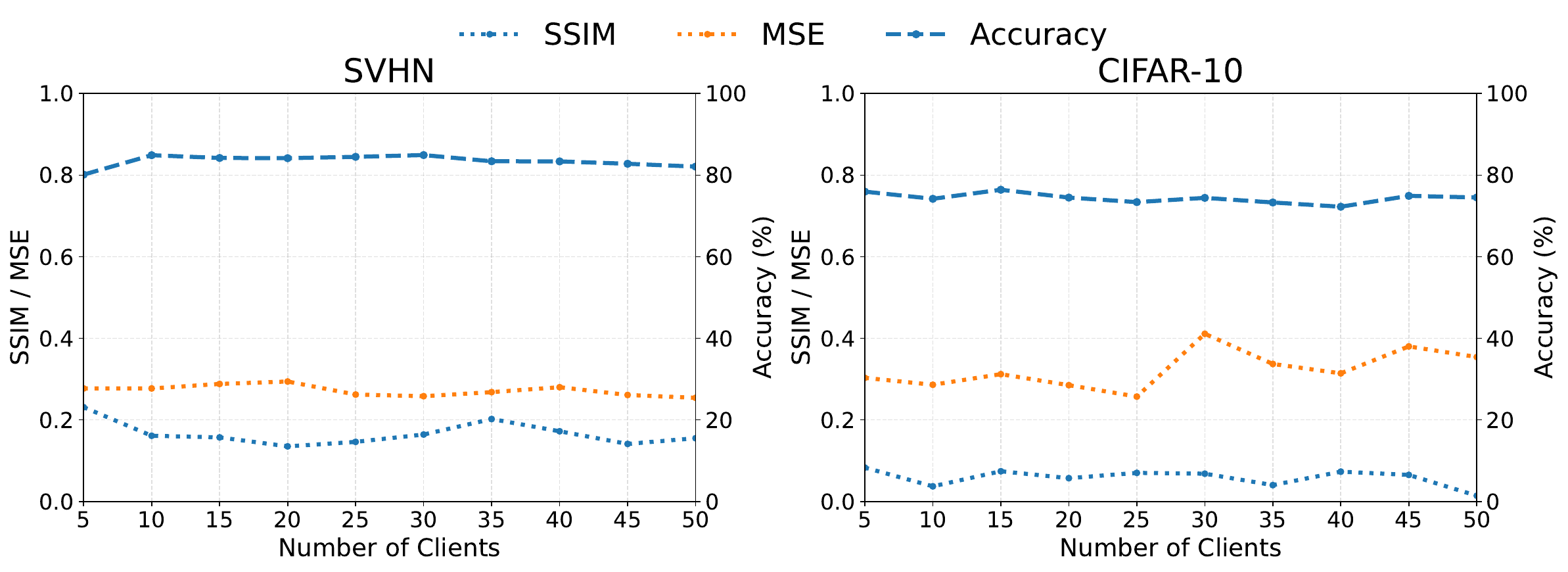}
    \caption{Impact of number of clients on SSIM, MSE, and accuracy in the KD-UFSL framework with the ConvNet architecture.}
    \label{fig:clients}
\end{figure}

This section investigates the impact of scale in the KD-UFSL framework. We compare the SSIM and MSE scores to evaluate the privacy of the framework and the accuracy to assess the utility of the framework as the number of clients increases from 5 to 50. As depicted in \textit{Figure} \ref{fig:clients}, it can be seen that the increase in the number of clients has a minimal effect on all three metrics in the framework, thus proving that the KD-UFSL framework can scale to a higher number of participants.
\section{Conclusion}

This work proposes k-anonymized differentially private federated split learning (KD-UFSL). This novel UFSL approach utilizes data-level differential privacy and k-anonymity, specifically micro-aggregation on the smashed data, to reduce the risk of data reconstruction from the aggregated data. Intermediate representations or smashed data can leak private client data in UFSL in the presence of an adversary, and KD-UFSL reduces the risk by adding noise to the data and anonymizing intermediate representations. The experiments indicate that KD-UFSL can increase the mean squared error between the actual and reconstructed data by up to 50\%, and reduce the image similarity, indicated by SSIM, by up to 40\%, compared to the vanilla UFSL setup, while preserving utility. In our experiments, the utility of the global model, measured by the global model accuracy, only drops by about 2-2.5\%.

\bibliography{main}

\end{document}